\documentclass[11pt]{article}
\usepackage{ranlp}
\usepackage{epsf} \newcommand{\opres}[0]{\approx}

\newcommand{\assign}[2]{[#1 = \textrm{#2}]}
\newcommand{\constraint}[3]{#1 $\opres^{#3}$ #2}

\newcommand{\rolevalue}[3]{{\sl (#1, #3, #2)}}
\newcommand{\varrole}[1]{{\sl #1}$_{\rm role}$}
\newcommand{\varmodel}[1]{{\sl #1}$_{\rm model}$}
\newcommand{\attval}[2]{#1:#2\\}
\newcommand{\negreta}[1]{\textbf{#1}}

\newcommand \modelGrupo[0]{group}

\newcommand \ModDePolicia[0]{\rolevalue{elem}{\ModDePolicia}{Policia}}
\newcommand \ModDeDroga[0]{\rolevalue{elem}{\ModDeDroga}{Droga}}

\newcommand{\SO}[1]{
\begin{tabular}{|l|}
#1
 \end{tabular}
}

\title{Integrating Multiple Knowledge Sources for Robust Semantic Parsing}

\author{Jordi Atserias,  Llu\'{\i}s Padr\'o \& German Rigau \\
        TALP Research Center \\
        Universitat Polit\`ecnica de Catalunya\\
        C/ Jordi Girona Salgado, 1-3. 08034 Barcelona, Catalonia \\
        \texttt{batalla,padro,g.rigau@lsi.upc.es} \\
        }

\begin{document}

\maketitle

\begin{abstract}
This work explores a new robust approach for
  Semantic Parsing of unrestricted texts. Our approach considers
  Semantic Parsing as a Consistent Labelling Problem ({\sc clp}), allowing
  the integration of several knowledge types (syntactic and semantic)
  obtained from different sources (linguistic and statistic). The
  current implementation obtains 95\% accuracy in model
  identification and 72\% in case-role filling. 
\end{abstract}

%%%%%%%%%%%%%%%%%%%%%%%%%%%%%%%%%%%%%%%%%%%%%%%%%%%%%%%%%%%%%
\section{Introduction}

A central issue in Semantic Parsing is the production of a
case-role analysis in which the semantic roles --such
as \emph{agent} or \emph{instrument}-- played by each entity 
are identified \cite{Brill+Mooney'97}.
This is a crucial task in any application which involves some level of
Natural Language Understanding.

This paper presents a new approach to Semantic Parsing ({\sc sp}). The aim of
this research is to develop a robust (able to work on unrestricted
text) and flexible (portable and extensible) approach to Semantic
Parsing. We will try to do so by formalizing semantic parsing as
an Consistent Labelling Problem ({\sc clp}), specially focusing on
the interaction between syntax and semantics, as well as on verbs,
as the head sentence components.

Mahesh \cite{Mahesh'93} proposes a classification of Natural Language
Understanding models in: \emph{sequential}, \emph {integrated} and
\emph{interactive} depending on the interaction between syntax and
semantics. 
In \emph{Sequential} models, each level receives the
output of the previous one, and sends its output to the next. Thus,
syntax is solved before any semantic analysis is carried out.  On the
other hand, in \emph{Integrated} and \emph{Interactive} models,
syntactic and semantic processing are performed simultaneously and
share a common knowledge representation.  

%The models differ in the following: while the \emph{Integrated} model
%(e.g. SAL \cite{Jurafsky'92}) uses a common representation for
%syntactic and semantic knowledge, the \emph{Interactive} model (e.g.
%COMPERE \cite{Eiselt+'93}) retains an independent representation of
%different kinds of knowledge and integrates them during the processing
%phase.

%A common intermediate approach between \emph{Sequential} and
%\emph{Interactive} models are \emph{Tandem} or \emph{Interleaved}
%systems, where the process alternates syntactic and semantic analysis.
%Further, these Tandem systems can be divided into those that are
%syntax driven and those that are semantics driven. The former entail
%the semantic validation of partial syntax results (e.g. ABSITY
%\cite{Hirst'87} and Hunter Gatherer \cite{Beale+Nirenburg'95}). In the
%latter, semantic attachments are proposed and grammar rules are
%searched among the constituents so as to accomplish these attachments
%(e.g. MOPTRANS/LINK \cite{Lytinen'86,Lytinen'91}).
%The main drawback of these approaches is that only one of both kinds of
%knowledge, syntax or semantics, drives the interpretation process, and
%that is a design decision that must be established in advance.

Another relevant issue related to syntax and semantics interaction is
the required level of syntax analysis.  Chunk parsing \cite{Abney'91}
has been widely used in several fields (e.g. Information Extraction)
as an alternative to deal with the lack of robustness presented by
traditional full parsing approaches: close world assumption (full
coverage grammar and lexicon), local errors produced by global parsing
considerations \cite{Grishman'95} and the selection of the best parse
tree among a forest of possible candidates.  
%Chunk parsing approach
%consists in building up small chunks using local syntactic criteria
%and then assembles larger structures only if they are semantically
%licensed.
Given that the verb is the central sentence component, there is no
doubt that subcategorization information may  not only improve
parsing ---e.g. taking into account probabilistic subcategorization on a
statistical parser \cite{Carroll+'98}, but also provide the basic 
information to assemble those chunks in larger structures.

Following this brief introduction,
sections~\ref{approach}, ~\ref{architecture} and ~\ref{formalization} present, respectively,
the basic ideas of our system and its architecture,
as well as the way in which the different sources of knowledge 
are integrated.
Section~\ref{experiments} describes the experiments
carried out and reports the results obtained. Finally,
section~\ref{f-work} presents some conclusions and 
outlines further research lines.

%%%%%%%%%%%%%%%%%%%%%%%%%%%%%%%%%%%%%%%%%%%%%%%%%%%%%%%%%%%%%
\section{{\sc pardon} approach}
\label{approach}

Our view of semantic parsing is based on compositional semantics and 
lexicalized models (i.e. the meaning of a sentence is the result
 of combining the meaning of its words and the possible combinations are
determined by their models).
Bearing that in mind, {\sc pardon} approach combines the
\emph{Interactive Model} and chunk parsing.  
Roughly speaking, {\sc pardon} combines semantic objects associated to chunks 
in order to build a case-role representation of the sentence.
This combination is carried out using syntactic and semantic knowledge
obtained from a linguistic approach (subcategorization frames)
and complemented with a statistical model of lexical attraction.

%-------------------------------------------------------------------------
\begin{figure*}[htb]\centering
\footnotesize{
\begin{tabular}{cccccc}
Este a\~no & en el congreso & del partido & se & habl\'o & de las pensiones \\
\\
\SO{
\attval{head}{a\~no}
\attval{hdle}{}
\attval{pos}{NP}
\attval{num}{sg} 
\attval{gen}{m}
\attval{sem}{Time} 
} &
\SO{
\attval{head}{congreso}
\attval{hdle}{en}
\attval{pos}{PP}
\attval{num}{sg} 
\attval{gen}{m}
\attval{sem}{Top} 
} &
\SO{
\attval{head}{partido}
\attval{hdle}{de}
\attval{pos}{PP}
\attval{num}{sg} 
\attval{gen}{m}
\attval{sem}{Group Human}
} &
\SO{
\attval{head}{se}
\attval{hdle}{}
\attval{pos}{se}
\attval{num}{} 
\attval{per}{}
\attval{sem}{Top}
} &
\SO{
\attval{head}{hablar}
\attval{hdle}{hablar}
\attval{pos}{VP}
\attval{num}{sg} 
\attval{per}{3}
\attval{sem}{Commun.}
} &
\SO{
\attval{head}{pensi\'on}
\attval{hdle}{de}
\attval{pos}{PP}
\attval{num}{pl} 
\attval{per}{3}
\attval{sem}{Top}
} \\
\end{tabular}
}
\caption{Chunks for ''Este a\~no en el congreso del partido se
habl\'o de las pensiones''}
\label{CS1} 
\end{figure*}

%-------------------------------------------------------------------------
\begin{figure*}[htb] \centering
\SO{
\SO{
\\Este a\~no\\
\\
\\
}
\SO{
\attval{model}{impersonal}
\attval{event}{habl\'o}
\attval{se}{se}
\attval{entity}{de las pensiones}
}
\SO{
\attval{model}{N$_{de}$}
\attval{head}{en el congreso}
\attval{modif}{del partido}
\\
}
}
\caption{Case-role structures obtained for the sentence in Figure~\ref{CS1}}
\label{result}
\end{figure*}

For instance,
starting from the chunks in the sentence ``Este a\~no en el congreso
del partido se habl\'o de las pensiones''\footnote{\emph{Literal
Translation: This year in the meeting of
the political party one talked about the pensions}} shown in
Figure \ref{CS1}, we will obtain the case-role representation shown
in Figure~\ref{result} by combining:
\begin{itemize}
\itemsep -0.15cm
\item The initial semantic objects associated to those chunks
\item The impersonal model of the verb ``hablar'' (to talk) shown in
  Table \ref{mHablar}
\item The noun modifier model shown in Table \ref{mNde}
\end{itemize}

\section{{\sc pardon} architecture}
\label{architecture}

We propose a novel architecture where high level syntax and semantics
decisions are fully integrated. 
There are two main steps in {\sc pardon}:
 The first step is the \emph{Sentence Analyzer}, which performs
 PoS-tagging, chunking and semantic annotation. It also accesses
the subcategorization (hand build) and lexical attraction
(statistical) knowledge bases, to complete the sentence model with
those kinds of knowledge.

 The verbal subcategorization information is obtained from LEXPIR
\cite{Fernandez+'96,Fernandez+'99,Morante+'98}
developed inside the Pirapides\footnote{Pirapides is centered on the
study of the English, Spanish and Catalan verbal predicates, based on
Levin \cite{Levin'93} and Pustejovsky's \cite{Pustejovsky'95}
theories.  LEXPIR models include information about the number of
arguments, their syntactic realization, prepositions they
can take, their semantic component and restrictions, their 
agreement with the verb and their optionality.} project.
Table \ref{mHablar} shows the basic and
impersonal models for the verb ``hablar'' (\emph{to talk}).  In this
work we used 61 verbs belonging to the LEXPIR trajectory class, which
includes the equivalent of Levin's movement and communication classes.

\begin{table*}[hbt] \centering
\begin{tabular}{|llllll|} \hline
\multicolumn{6}{|c|}{{\sl basic} model for ``hablar''} \\
 Synt. & Prep. & Comp. & Seman. & Agree. & Opt. \\ \hline
NP  & x          & starter     & Human & yes & yes \\ 
PP  & de, sobre  & entity      & Top   & no  & yes \\ 
PP  & con        & destination & Top   & no  & yes \\ \hline 
\multicolumn{6}{c}{}  \\ \hline
\multicolumn{6}{|c|}{{\sl impersonal} model for ``hablar''}  \\
 Synt. & Prep. & Comp. & Seman. & Agree. & Opt. \\ \hline
SE  & x          & se          & Top   & no  & no \\ 
PP  & de, sobre  & entity      & Top   & no &  yes \\ 
PP  & con        & destination & Top   & no & yes \\ \hline 
\end{tabular}
\caption{Models for the verb ``hablar''}
\label{mHablar}
\end{table*}

  The second step is \emph{Selection}, which solves the Consistent 
Labelling Problem associated to that sentence model to find out which 
is the most appropriate role for each chunk.

\section{{\sc pardon} Formalization}
\label{formalization}

We formalize {\sc pardon} approach by setting the case-role
interpretation problem as a \emph{Consistent Labelling Problem} 
({\sc clp}), where the different kinds of knowledge are 
applied as weighted constraints.

A {\sc clp} basically consists of finding the most consistent label assignment
for a set of variables, given a set of constraints. 
Once the sentence and its knowledge is represented in terms of a {\sc clp},
a relaxation labelling algorithm is used to obtain the most consistent
interpretation. See \cite{Padro'98} for details on the use of
these algorithm for {\sc nlp} tasks.

This formulation allows us to naturally integrate different kinds of knowledge
coming from different sources (linguistic and statistic), which
may be partial, partially incorrect or even inconsistent.

%%%%%%%%%%%%%%%%%%%%%%%%%%%%%%%%%%%%%%%%%%%%%%%%%%%%%%%%%%%%%
%\section{{\sc pardon} architecture}
%\label{architecture}

{\sc pardon} represents the meaning of a sentence in terms of relationships
between semantic objects, using two variables for each semantic
object: the \emph{model} (\varmodel{object}) and
\emph{role} (\varrole{object}) variables. 
For instance, the semantic object associated to a chunk headed by
``hablar'' (to talk) can use a \emph{basic} model (someone talks about
something with someone: \assign{\varmodel{hablar}}{\emph{basic}}) or an
\emph{impersonal} model (\emph{one} talks about something \assign{\varmodel{hablar}}{\emph{impersonal}}).

The \emph{role} variable represents the role that a
semantic object plays inside the model of another semantic object. For
instance, the semantic object ``pensiones'' (the pensions) can play
the role \emph{entity} for both models of ``hablar'' (to talk)
(e.g. \assign{\varrole{pensiones}}{\rolevalue{entity}{hablar}{basic}}) 

To identify a role from a model label we need a triple
\rolevalue{role}{semantic object}{model}. For instance, the role 
\emph{starter} of the \emph{basic} model for
``hablar'', represented as \rolevalue{starter}{hablar}{basic}.

Since a {\sc clp} always assigns a label to all the variables, two null
labels have to be added: the label {\sc none} for the model variables
(semantic objects which does not have/use a model, usually leaf
semantic objects with no sub-constituents) and the label {\sc top}
for the role variables (semantic objects not playing a role
in the model of a higher constituent, usually the sentence head).
Figure \ref{VL1} shows the variables and labels associated to the
semantic objects in Figure \ref{CS1}.

%---------------------------------------------------------------
\begin{figure}[hbt]
\centering
\begin{tabular}{|l|l|}
      \emph{Variable Name}  & \emph{Possible Labels}  \\ \hline \hline
        \varmodel{a\~no} & \modelGrupo \\ 
                              & NONE \\ \hline
        \varrole{a\~no}  &\rolevalue{starter}{hablar}{basic} \\
                            & TOP \\ \hline \hline

         \varmodel{congreso} & \modelGrupo \\ 
                              & NONE \\ \hline
         \varrole{congreso} & TOP \\ \hline \hline

         \varmodel{partido} & \modelGrupo  \\ 
                        & NONE \\ \hline
         \varrole{partido}  & \rolevalue{entity}{hablar}{basic} \\ 
                             & \rolevalue{entity}{hablar}{impersonal}\\
                             & \rolevalue{modif}{a\~no}{N$_{de}$} \\
                             &\rolevalue{modif}{congreso}{N$_{de}$} \\ 
                             &\rolevalue{modif}{pension}{N$_{de}$} \\ 
                             & TOP \\ \hline \hline

         \varmodel{se}  & NONE \\ \hline 
         \varrole{se}   &       \rolevalue{se}{hablar}{impersonal}  \\ 
                                & TOP \\ \hline \hline
         \varmodel{hablar}          & basic   \\
                                    & impersonal   \\ 
                             & NONE \\ \hline
         \varrole{hablar} & TOP \\ \hline \hline

         \varmodel{pensi\'on} & \modelGrupo  \\ 
                        & NONE \\ \hline
         \varrole{pensi\'on} & \rolevalue{entity}{hablar}{basic} \\ 
                            & \rolevalue{entity}{hablar}{impersonal}\\ 
                            & \rolevalue{modif}{a\~no}{N$_{de}$} \\ 
                            &\rolevalue{modif}{congreso}{N$_{de}$} \\ 
                            &\rolevalue{modif}{partido}{N$_{de}$} \\ 
                        & TOP \\ \hline
\end{tabular}
\caption{{\sc clp} associated to the objects in Figure~\ref{CS1}}
\label{VL1}
\end{figure}

After formalizing Semantic Parsing as a Consistent Labelling
Problem, a set of constraints stating valid/invalid 
assignations is required to find the solution. 
{\sc pardon} uses three kinds of constraints: The first group contains the
constraints that encode the linguistic information obtained from verb
subcategorization models. The second group are additional constraints
added to force a tree-like structure for the solution. Finally, a
third set of constraints encoding statistical information about word
co\-occurrences, was added in order to complement the
subcategorization information available.

 Constraints are noted as follows: 

\constraint{\assign{A}{x}}{\assign{B}{y}}{w} 
%
%denotes a constraint stating that variable $A$ having label $x$ has a compatibility
%degree $w$ with variable $B$ having label $y$. 
%
denotes a constraint stating a compatibility degree $w$ when variable
$A$ has label $x$ and variable $B$ has label $y$. The compatibility
degree $w$ may be positive (stating compatibility) or negative
(stating incompatibility).

\subsection{Subcategorization Constraints}
\label{LEXPIR}

Two different kinds of subcategorization models have been used: one
about verbal subcategorization and one about noun modifiers.

For each chunk labelled as VP, all possible subcategorization 
models for the verb heading the chunk are retrieved from LEXPIR. 
For PP and NP we use the simple nominal modifier model N$_{de}$ 
presented in table~\ref{mNde}. 

\begin{table*}[hbt] \centering
\begin{tabular}{|llllll|} \hline
\multicolumn{6}{|c|}{{\sl N$_{de}$} model for nouns}  \\
 Synt. & Prep. & Comp. & Seman. & Agree. & Opt. \\ \hline
PP      & de    & modifier  & Top       & no     &  no \\ \hline 
\end{tabular}
\caption{Model for noun modifiers}
\label{mNde}
\end{table*}

Due to the richness of natural language
we cannot expect to find, in a real sentence sample, the exact
prototypical subcategorization patterns that have been modelled in
LEXPIR. Thus, a measure of the "goodness" of the possible model
instantiation is defined in a similar way to the tree-edit based
pattern matching used in \cite{Atserias+'99,Atserias+'00}.

In order to ensure the global applicability
(minimal disorder, agreement, maximum similarity between the role and
semantic object and maximal number of roles) and the consistence of
the model (a unique instantiation per role and the instantiation of 
compulsory roles) the following constraints are automatically instantiated
from the models:
\begin{itemize}
\itemsep -0.1cm
\item \negreta{Role Uniqueness}: The same role can not be assigned to different
chunks, e.g.:

\constraint{\assign{\varrole{pension}}{\rolevalue{entity}{hablar}{basic}}}
           {\assign{\varrole{partido}}{\rolevalue{entity}{hablar}{basic}}}
           {-1}

This constraint penalizes the current weight of the assignment
\assign{\varrole{pension}}{\rolevalue{entity}{hablar}{basic}}
according to
the current weight of the assignment \assign{\varrole{partido}}{\rolevalue{entity}{hablar}{basic}}.
Thus, the higher the weight for the latter assignment is, the faster 
the weight of the former will decrease.
\item \negreta{Model Support}: A model assignment is compatible with
its optional roles, e.g.:

\constraint{\assign{\varmodel{hablar}}{basic}}
           {\\ \assign{\varrole{pension}}{\rolevalue{entity}{hablar}{basic}}}
           {+1}
\item \negreta{Model Inconsistence}: A model assignment is incompatible
with the inexistence of any of its compulsory roles, e.g.:

\constraint{\assign{\varmodel{hablar}}{impersonal}}
           {\\ $\neg$ \assign{\varrole{se}}{\rolevalue{se}{hablar}{impersonal}}}
           {-1}
\item \negreta{Role Support}: A role assignment is compatible with the
assignment of its model, e.g.:

\constraint{\assign{\varrole{pension}}{\rolevalue{entity}{hablar}{basic}}\\}
           {\\ \assign{\varmodel{hablar}}{basic}}
           {+sim(pension,\rolevalue{entity}{hablar}{basic})}

The weight for this constraint is defined as a function $sim$, which
measures the similarity between two feature structures yielding
a value normalized in $[-1,1]$, inversely proportional to the number of 
relabelling operations needed to transform one feature
structure into the other. Currently, only semantics, gender
and number are considered.
\item \negreta{Role Inconsistence}: A role assignment is incompatible
with the \emph{no existence} of the assignment of its own  model, e.g.:

\constraint{\assign{\varrole{pension}}{\rolevalue{entity}{hablar}{basic}}}
           {\\ $\neg$ \assign{\varmodel{hablar}}{basic}}
           {-1}
\end{itemize}
 
Additionally, a special set of constraints has been introduced to deal with PP-attachment:
\begin{itemize}
\itemsep -0.1cm
\item \negreta{Local PP attachment}: A prepositional phrase tends to
  be attached to its nearest head. The weight assigned to each
  constraint will decrease along with the distance (in words) between
  the semantic objects involved, e.g.:

\constraint{\assign{\varrole{pension}}{\rolevalue{entity}{hablar}{impersonal}}}
           {[ ]}
           {-distance(pension,hablar)}.
\end{itemize}

\subsection{Structural Constraints}

Some further constraints must be included to force the solution to
have a tree-like structure. These constraints are not derived from
the subcategorization models.

\begin{itemize}
\itemsep -0.1cm
\item \negreta{TOP Uniqueness}: Different assignments of the label TOP
are incompatible, e.g.:

\constraint{\assign{\varrole{partido}}{TOP}}
           {\assign{\varrole{hablar}}{TOP}}
           {-1}.
\item \negreta{TOP Existence}: There is at least a TOP. Notice that
there is no right side on the constraint as it is valid for any context, e.g.:

\constraint{\assign{\varrole{hablar}}{TOP}}{[ ]}{+1}
\item \negreta{No Cycles}: Two assignments forming a direct cycle
are incompatible\footnote{In this first prototype of \emph{{\sc pardon}} indirect cycles are not taken into account}, e.g.:

\constraint{\assign{\varrole{pension}}{\rolevalue{modif}{partido}{N$_{de}$}}}
           {\assign{\varrole{partido}}{\rolevalue{modif}{pension}{N$_{de}$}}}
           {-1}
\item \negreta{NONE Support}: The NONE model is compatible with the
  inexistence of any role assignment of the semantic object models,
  e.g.:

\constraint{\assign{\varmodel{congreso}}{NONE}}
    {\\$\neg$ \assign{\varrole{pension}}{\rolevalue{modif}{congreso}{N$_{de}$}} $\wedge$\\
     $\neg$ \assign{\varrole{partido}}{\rolevalue{modif}{congreso}{N$_{de}$}}}
    {+1}

%
%  If this constraint were not included, the NONE model would never receive any support, and would
% never be selected, since there would always be some other model with a tiny non-zero support.
%
If these constraints were not included, the NONE model would never
be selected, since there would always be
some other model with a very small non-zero support.
\end{itemize}

\subsection{Statistical Constraints}
\label{lexatt}

In a similar way to \cite{Yuret'98} we define also a language
model based on lexical attraction. In our case, we estimate
the likelihood of a syntactic relation not between two words but between
two semantic objects.

Our hypothesis is that the relations between two semantic objects can be
determined taking into account two special elements of their associated chunks, the \emph{handle} and the \emph{head}.
The \emph{handle} of a chunk is usually the preposition which 
specifies the type of relation it has with another chunk, while 
the \emph{head} of a chunk is
supposed to capture the meaning of the chunk \cite{Basili+'98}. 
For instance, the chunk ``de las pensiones'' (\emph{about the pensions}) has 
handle ``de'' (\emph{about}) and \emph{head} ``pensi\'on'' (\emph{pension}).

Since related words are expected to occur together more likely than 
unrelated words,
the lexical attraction (the likelihood of a syntactic relation) between
two words can be estimated/modeled through co\-ocurrence. Co\-occurence
data can also indicate negative relatedness, where the probability of
co\-occurence is less than by chance. 
Thus, we will measure lexical attraction between two sematic objects 
with the
co\-ocurrence of both heads and the co\-occurrence of the head and the
handle (which gives an implicit
direction of the dependence). 
 
Since the co\-ocurrences were taken from the definitions of a Spanish
dictionary, lemma co\-ocurrences were used instead of word co\-ocurrences in order to
minimize the problems caused by unseen words \cite{Dagan+'99}. 175,333
head-handle co\-ocurrences and 961,470 head-head co\-occurences were
obtained out of 40,591 different head-lemmas and 160 different
handle-prepositions. The co\-occurrences were used to compute Mutual Information for each lemma--preposition pair.
$$
MI(head_i,handle_j) = log \frac{P(head_i \cap handle_j)}{P(head_i)
  \times P(handle_j)} 
$$
%
%For lemma--lemma pairs sparseness is much higher, thus we used an indirect
%measure, such as context vector cosine --used in IR and WSD \cite{Schutze'92}-- as the lexical attraction between heads.
%
In the case of lemma-lemma pairs, sparseness is much higher. Thus, an
indirect measure was applied, namely context vector cosine (also used
in IR and WSD (Schütze, 1992)) in order to calculate the lexical
attraction between heads:
$$
cos(head_i,head_j) = \frac{\sum_{k} a_{ki} a_{kj}}{\sqrt{\sum_{k}
    {a}^2_{ki} \sum_{k} {a}^2_{kj}}}
$$ 
where $a_{pq}$ is the co\-occurrence frequency of lemma $p$ and lemma
$q$, and  $k$ ranges over all the lemmas co\-occurring with any of both heads.

Thus, for any two semantic objects the following constraints are added:

\begin{itemize}
\itemsep -0.1cm
\item $A_i$-$H_j$ constraint, which supports any assignment of a role
from $object_j$ to $object_i$, e.g.:

\constraint{\assign{\varrole{partido}}{\rolevalue{modif}{congreso}{N$_{de}$}}\\}
           {[ ]}
           {MI(congreso,de)}
\item $H_i$-$H_j$ constraint, which supports any assignment of a role from
$object_i$ to $object_j$, or viceversa, e.g.:

\constraint{\assign{\varrole{pensi\'on}}{\rolevalue{entity}{hablar}{impersonal}}\\}
           {[ ]}
           {cos(hablar,pension)}

\end{itemize}
$H_i$-$H_j$ and $A_i$-$H_j$ constraints can be used to identify adjuncts
or relations for which we have no models. For instance, in the result 
obtained for the sentence shown in Figure \ref{CS1}, the sematic object ``en 
el congreso'' (\emph{in the meeting}) will be identified as depending on the 
verb ``hablar'', even when its role can not be determined.  

\section{Experiments}
\label{experiments}

170 real sentences were taken from a Spanish newspaper
and were labelled by hand with the verbal models and the meaning
components. The sentence average length is 8.1 words, ranging
from 3 to 23. Only one-verb sentences were selected, since 
our knowledge base does not include models for subordination 
or coordination. However, our approach to semantic parsing has been
designed to manage multiple models simultaneously competing for their
arguments. 

Each sentence in the corpus was tagged and parsed with a wide-coverage
grammar of Spanish~\cite{Castellon+'98} to obtain a 
chunk parse tree. Spanish Wordnet \cite{Atserias+'97}
was used to semantically annotate the corpus with the 79 semantic
labels defined in the preliminary version of the EuroWordnet Top
Ontology \cite{Rodriguez+'98}. 

In order to reduce the complexity of the relaxation process, the
possible role labels (which indicate the roles an object can play in
any of the models retrieved) are filtered considering the unary
constraints about POS and prepositions, while
constraints about semantics and agreement are taken as a measure of
how similar ($sim$) is the semantic object and the role. Models which
can not match compulsory roles are not considered.

For instance, the
semantic object \emph{a\~no} (\emph{year}) in the example sentence
will be allowed to match the role \emph{starter} of the impersonal
model of the verb \emph{hablar} even though its semantics is not
Human, but the semantic object \emph{congreso} will not be
considered as a candidate to fill the \emph{entity} role of
\emph{hablar}, since the preposition \emph{en} in the semantic object
does not match the model requirements for that role (preposition
\emph{de, sobre}). 

All these filters produce the candidate labels
shown in Figure~\ref{VL1}, and are the input to {\sc pardon}
\emph{Selection} step.

\subsection{Results}

%!!!  Jordi: es MUC-6 o MUC-7 ??
The results reported have been calculated using Message Understanding
Conferences \cite{MUC6} evaluation metrics applied to our
particular case of verbal model identification and case-role
filling. 

Model identification metrics evaluate how well our system identifies
the right model for a semantic object. Our corpus has 2.7 models per
verbal semantic object as average ambiguity.

Since it is assumed that there is only one right model per chunk in
each sentence, the answer can only be correct ($COR$) or incorrect
($INC$), thus, the used metrics are precision and recall.  Table
\ref{resmodel} shows the results obtained in the verbal model
identification task: 95\% precision and 91\% recall.

\begin{table}[htb] \centering
\begin{tabular}{|c|c||c|c|}
  $COR$ & $INC$ & $PRE$ & $REC$ \\ \hline
  155   &    8  &  95\% & 91\% \\ \hline
\end{tabular}
\caption{Verbal Model identification results}
\label{resmodel}
\end{table}

Case-role filling consists in assigning each semantic object to the
right role it plays in the models for other semantic objects. In this
case, the casuistics is more complex, since in addition to the
correct/incorrect distinctions, other cases must be considered, such as
the roles that are (correctly/incorrectly) left unassigned (because
they were optional, or there was no semantic object that fitted them,
etc.). The MUC evaluation metrics establish the following cases:
\begin{itemize}  
\itemsep -0.15cm
\item \negreta{Correct} ($COR$): Roles correctly assigned by the system.
\item \negreta{Incorrect} ($INC$): Roles incorrectly assigned by the system.
\item \negreta{Missing} ($MIS$): Roles unassigned by the system when they should have been assigned.
\item \negreta{Spurious} ($SPU$): Roles assigned by the system when they should have been unassigned.
%\item \negreta{Possible} ($POS$): Roles that should be assigned
%  plus optional roles that should be assigned and that the system generates.
%\item \negreta{Actual} ($ACT$): Roles actually assigned by the system.
\end{itemize}
These cases lead to the definition of the following measures, where \negreta{Possible} ({\sc pos}) are the roles that should be assigned
({\sc cor}+{\sc inc}+{\sc mis}) and \negreta{Actual} ({\sc act}) are
the roles actually assigned by the system under evaluation ({\sc
cor}+{\sc inc}+{\sc spu}):
\begin{itemize}
\itemsep -0.1cm
\item \negreta{Undergeneration} $UND = 100\times\frac{MIS}{POS}$
\item \negreta{Overgeneration}  $OVR = 100\times\frac{SPU}{ACT}$
\item \negreta{Substitution}    $SUB = 100\times\frac{INC}{COR+INC}$
\item \negreta{Error}      $ERR = 100\times\frac{INC+SPU+MIS}{COR+INC+SPU+MIS}$
\item \negreta{Precision}  $PRE = 100\times\frac{COR}{ACT}$
\item \negreta{Recall}     $REC = 100\times\frac{COR}{POS}$
\end{itemize}
  In addition, precision and recall may be combined in different 
F-measures ($P\&R$, $2P\&R$ and $P\&2R$).
 Table \ref{resrole} shows the results in the
case-role filling for verbal arguments.
\begin{table}[hbt] \centering
\begin{tabular}{|c|c|c|c|c|c|} 
 $COR$ & $INC$ & $MIS$ & $SPU$ & $POS$ & $ACT$ \\ \hline
 203 & 27 & 60 & 51 & 290 & 281 \\ \hline
\end{tabular}
\begin{tabular}{|c|c|c|c|c|c|} 
\multicolumn{6}{c}{} \\ 
 $UND$ & $OVR$ & $SUB$ & $ERR$ & $PRE$ & $REC$ \\ \hline
20\% & 18\% & 12\% & 40\% & 72\% & 70\% \\ \hline
\end{tabular}
\begin{tabular}{|c|c|c|}
\multicolumn{3}{c}{} \\ 
 $P\&R$ & $2P\&R$ & $P\&2R$ \\ \hline
 71\% & 70\% & 72\% \\ \hline
\end{tabular}
\caption{Verbal case-role filling results}
\label{resrole}
\end{table}

To our knowledge there is neither a similar general approach nor 
case-role filling experiments to which our results can be compared. 
In any case, our preliminary results (72\% $PRE$ - 70\% $REC$) are
very encouraging.

% Aixo queda fatal. Si no es comparable, no ho comparem.
%However, although the results of our approach for sentence
%understanding and the MUC template element task (TE) on a specific
%domain are hardly comparable, our role filling precision and
%recall (72\% $PRE$ - 70\% $REC$) stand up to the results obtained
% by the different systems 
%in MUC-6 \cite{MUC6}, which ranged from (60\% $PRE$ - 49\% $REC$) to 
%(85\% $PRE$ - 71\% $REC$). 
%That is, we obtained for a non specific domain 
%similar performances than those obtained on MUC-6 for 
%restricted domains.

It is also remarkable that our system produces low values for
$UND$, $OVR$ and $SUB$ measures, pointing that it properly
uses the different kinds of knowledge, and that it does not take uninformed or 
gratuitous decisions.

 Errors in the preprocessing steps caused most of miss-identified models (table \ref{resmodel}, $INC$)
The {\sl missing} and {\sl spurious} roles (table \ref{resrole}, $MIS$ and
$SPU$) were due either to the lack of semantic information or to the 
lack of a verbal model for adjuncts, which
caused miss-identification of adjuncts as arguments, as in 
``(Juan) (esqu\'{\i}a) (este fin) (de a\~no)''\footnote{\emph{John
    goes skying on New Year's Eve}}, where the
chunk ``este fin de a\~no'' (\emph{on New Year's Eve}) is wrongly
identified to fill the \emph{route} role even though its semantics is
\emph{Time}.
This is due to the lack of a selectional restriction that forces
the \emph{route} to be a \emph{Place}, and to the lack of a model
that identifies the chunk as time adjunct.

%-----------------------------------------------------------------------------
\section{Conclusions \& Further Work}
\label{f-work}
This paper has presented a new approach to Semantic Parsing for non
domain-specific texts based on the \emph{Interactive Model}. 
The robustness and flexibility of {\sc pardon} is achieved
combining a chunk parsing approach with the framing of the semantic
parsing problem in a {\sc clp}.
The flexibility of our approach enables the integration of different
types of knowledge (linguistically motivated subcategorization models 
 plus statistical information obtained from corpora).

 Currently, {\sc pardon} obtains a 95\% precision on model 
identification and 72\% precision on role filling. Although the experiments
have been carried out on a limited corpus and lexicon, they have
proven the feasibility of the method.

Further work should approach a more realistic evaluation of the system,
using a larger corpus with multiple-verb sentences. In this
case, verbs will compete in a sentence for their arguments. We also 
plan to include more statistical knowlege
(measures/language models) and to extend the coverage and
expressiveness of the subcategorization models. 
Furthermore, the output of the current system could also be 
used as feedback to improve the existing verbal models.

Exploration of linguistic and statistical models for the
identification/distinction of verbal adjuncts should also be 
addressed, since it seems one of the
main causes of the miss-identification of the verbal arguments.

\section{Acknowledgements}
{\small
This work has been partially funded by the EU (IST-1999-12392),
and by the Spanish (TIC2000-0335-C03-02, TIC2000-1735-C02-02) and Catalan
(1997-SGR-00051) Governments.
}

\bibliographystyle{ranlp}
\begin{scriptsize}
\bibliography{jab}

\begin{thebibliography}{Fern\'andez \& Mart\'{\i} 96}
\itemsep 0.08cm

\bibitem[Abney 91]{Abney'91}
(Abney 91)
Steven Abney.
\newblock {\em Parsing by chunks}.
\newblock Kluwer Academic Publishers, 1991.

\bibitem[Atserias {\it et al.} 97]{Atserias+'97}
(Atserias {\it et al.} 97)
Jordi Atserias, Salvador Climent, Xavier Farreres, German Rigau, and Horacio
  Rodr\'{\i}guez.
\newblock Combining multiple methods for the automatic construction of
  multilingual wordnets.
\newblock In {\em Procceeding of RANLP'97}, pages 143--149, Bulgaria, 1997.
\newblock Also to appear in a Book.

\bibitem[Atserias {\it et al.} 99]{Atserias+'99}
(Atserias {\it et al.} 99)
J.~Atserias, I.~Castell\'on, M.~Civit, and G.~Rigau.
\newblock Using diathesis for semantic parsing.
\newblock In {\em Proceedings of Venecia per il Trattamento automatico delle
  lingue (VEXTAL)}, pages 385--392, Venecia, Italy, 1999.

\bibitem[Atserias {\it et al.} 00]{Atserias+'00}
(Atserias {\it et al.} 00)
J.~Atserias, I.~Castell\'on, M.~Civit, and G.~Rigau.
\newblock Semantic analysis based on verbal subcategorization.
\newblock In {\em Proceedings of the Conference on Intelligent Text Processing
  and Computational Linguistics (CICLing)}, pages 330--340, Mexico City,
  Mexico, 2000.

\bibitem[Basili {\it et al.} 98]{Basili+'98}
(Basili {\it et al.} 98)
R.~Basili, M.~T. Pazienza, and F.~Zanzotto.
\newblock Efficient parsing for information extraction.
\newblock In Henri Prade, editor, {\em Proceedings of the 13th European
  Conference on Artificial Intelligence ({ECAI}-98)}, pages 135--139,
  Chichester, August~23-28 1998. John Wiley \& Sons Ltd.

\bibitem[Brill \& Mooney 97]{Brill+Mooney'97}
(Brill \& Mooney 97)
Eric Brill and Raymond~J. Mooney.
\newblock {An Overview of Empirical Natural Language Processing}.
\newblock {\em {Artificial Intelligence Magazine}}, 18(14):13--24, Winter 1997.
\newblock Special Issue on Empirical Natural Language Processing.

\bibitem[Carroll {\it et al.} 98]{Carroll+'98}
(Carroll {\it et al.} 98)
John Carroll, G.~Minnen, and E.~Briscoe.
\newblock Can subcategorisation probabilities help a statistical parser?
\newblock In {\em Proceedings of the 6th ACL/SIGDAT Workshop on Very Large
  Corpora}, pages 118--126, Montreal, Canada, 1998.

\bibitem[Castell\'on {\it et al.} 98]{Castellon+'98}
(Castell\'on {\it et al.} 98)
Irene Castell\'on, Montse Civit, and Jordi Atserias.
\newblock Syntactic parsing of spanish unrestricted text.
\newblock In {\em Proceedings of the 1th Conference on Language Resources and
  Evaluation (LREC'98)}, Granada. Spain, 1998.

\bibitem[Dagan {\it et al.} 99]{Dagan+'99}
(Dagan {\it et al.} 99)
Ido Dagan, Lillian Lee, and Fernando C.~N. Pereira.
\newblock Similarity-based models of word coocurrence probabilities.
\newblock {\em Machine Learning}, (34):43--69, 1999.

\bibitem[Fern\'andez \& Mart\'{\i} 96]{Fernandez+'96}
(Fern\'andez \& Mart\'{\i} 96)
A.~Fern\'andez and M.~A. Mart\'{\i}.
\newblock Classification of psycological verbs.
\newblock {\em SEPLN}, (20), 1996.

\bibitem[Fern\'andez {\it et al.} 99]{Fernandez+'99}
(Fern\'andez {\it et al.} 99)
A.~Fern\'andez, M.~A. Mart\'{\i}, G.~V\'azquez, and I.~Castell\'on.
\newblock Establising semantic oppositions for typification of predicates.
\newblock {\em Language Design}, (2), 1999.

\bibitem[Grishman 95]{Grishman'95}
(Grishman 95)
Ralph Grishman.
\newblock Nyu system or where's the syntax?
\newblock In {\em MUC-6}, pages 167--175, 1995.

\bibitem[Levin 93]{Levin'93}
(Levin 93)
Beth Levin.
\newblock {\em English Verb Classes and Alterations: A preliminar
  Investigation}.
\newblock The University of Chicago Press, 1993.

\bibitem[Mahesh 93]{Mahesh'93}
(Mahesh 93)
Kavi Mahesh.
\newblock A theory of interaction and independence in sentence understanding.
\newblock Unpublished M.Sc. thesis, Georgia Institute of Technology, 1993.

\bibitem[Morante {\it et al.} 98]{Morante+'98}
(Morante {\it et al.} 98)
R.~Morante, Irene Castell\'on, and Gloria V\'azquez.
\newblock Los verbos de trayectoria.
\newblock In {\em Proceedings of the conference of the SEPLN}, 1998.

\bibitem[MUC95]{MUC6}
(MUC95)
{\em Sixth Message Undestanding Conference (MUC-6)}, 1995.
\newblock ISBN 1-55860-402-2.

\bibitem[Padr\'o 98]{Padro'98}
(Padr\'o 98)
Lluis Padr\'o.
\newblock {\em A Hybrid Environment for Syntax-Semantic Tagging}.
\newblock PhD thesis, Universitat Politectnica de Catalunya, 1998.

\bibitem[Pustejovsky 95]{Pustejovsky'95}
(Pustejovsky 95)
James Pustejovsky.
\newblock {\em The generative lexicon}.
\newblock The MIT Press, Cambridge, 1995.

\bibitem[Rodr\'{\i}guez {\it et al.} 98]{Rodriguez+'98}
(Rodr\'{\i}guez {\it et al.} 98)
Horacio Rodr\'{\i}guez, Salvador Climent, Peek Vossen, L.~Blocsma, Wim Peters,
  A.~Alonge, F.~Bertagna, and A.~Rovertini.
\newblock The top-down strategy for building euwn: Vocabulary coverage, base
  concepts and top ontology.
\newblock {\em Computers and the Humanities}, 32(2-3), 1998.

\bibitem[Yuret 98]{Yuret'98}
(Yuret 98)
Deniz Yuret.
\newblock {\em Discovery of Linguistic Relations Using Lexical Attraction}.
\newblock Unpublished PhD thesis, Massachusetts Institute of Technology, 1998.
\newblock Also availabel as cmp-lg/9805009.

\end{thebibliography}
\end{scriptsize}
\end{document}